\documentclass[letterpaper, 10 pt, conference]{ieeeconf}

\IEEEoverridecommandlockouts                              
                                                          
\usepackage{tikz}
\usetikzlibrary{tikzmark}

\usepackage{cite}

\usepackage{amsmath}
\usepackage{algorithm}
\usepackage{algpseudocode}

   \usepackage{graphicx}
   \usepackage{svg}
   \usepackage{float}

\usepackage[T1]{fontenc}

\usepackage{array}

\usepackage{amsmath}
\usepackage{mathtools}
\usepackage{amssymb}
\usepackage{breqn}

  \usepackage[caption=false,font=normalsize,labelfont=sf,textfont=sf]{subfig}
\usepackage{fixltx2e}

\usepackage{stfloats}
\usepackage{url}

\usepackage{algorithm}
\usepackage{algpseudocode}
\usepackage{multirow}
\usepackage{mathtools}
\usepackage{cuted}

\usepackage[utf8]{inputenc}
\usepackage[T1]{fontenc}
 \usepackage{hyperref}
\begin{document}
%
\title{\LARGE \bf Semantic Enrichment of CAD-Based Industrial Environments \\ via Scene Graphs for Simulation and Reasoning}


\author{
Nathan Pascal Walus$^{1,3}$,Ranulfo Bezerra$^{1,2}$, Shotaro Kojima$^{1,2}$, Tsige Tadesse Alemayoh$^{1,2}$,\\ Satoshi Tadokoro$^{1,2}$, Kazunori Ohno$^{1,2}$
\thanks{*This research was performed by the commissioned research fund provided by F-REI (JPFR23010101).}
\thanks{$^{1}$Graduate School of Information Sciences, Tohoku University, Japan.}
\thanks{$^{2}$Tough Cyberphysical AI Research Center, Tohoku University, Japan.
}
\thanks{$^{3}$RWTH Aachen University, Aachen, Germany.}
\thanks{\tt\small bezerra.ranulfo@tr.is.tohoku.ac.jp}
\thanks{\tt\small nathan.walus@rwth-aachen.de}
}


%


\maketitle

\begin{abstract}
Utilizing functional elements in an industrial environment, such as displays and interactive valves, provide effective possibilities for robot training. When preparing simulations for robots or applications that involve high-level scene understanding, the simulation environment must be equally detailed. Although CAD files for such environments deliver an exact description of the geometry and visuals, they usually lack semantic, relational and functional information, thus limiting the simulation and training possibilities. A 3D scene graph can organize semantic, spatial and functional information by enriching the environment through a Large Vision-Language Model (LVLM). In this paper we present an offline approach to creating detailed 3D scene graphs from CAD environments. This will serve as a foundation to include the relations of functional and actionable elements, which then can be used for dynamic simulation and reasoning. Key results of this research include both quantitative results of the generated semantic labels as well as qualitative results of the scene graph, especially in hindsight of pipe structures and identified functional relations. All code, results and the environment will be made available at \url{https://cad-scenegraph.github.io}.
\end{abstract}
%
\IEEEpeerreviewmaketitle

\section{Introduction}


\begin{figure*}[b]
    \centering
    \includegraphics[width=\linewidth]{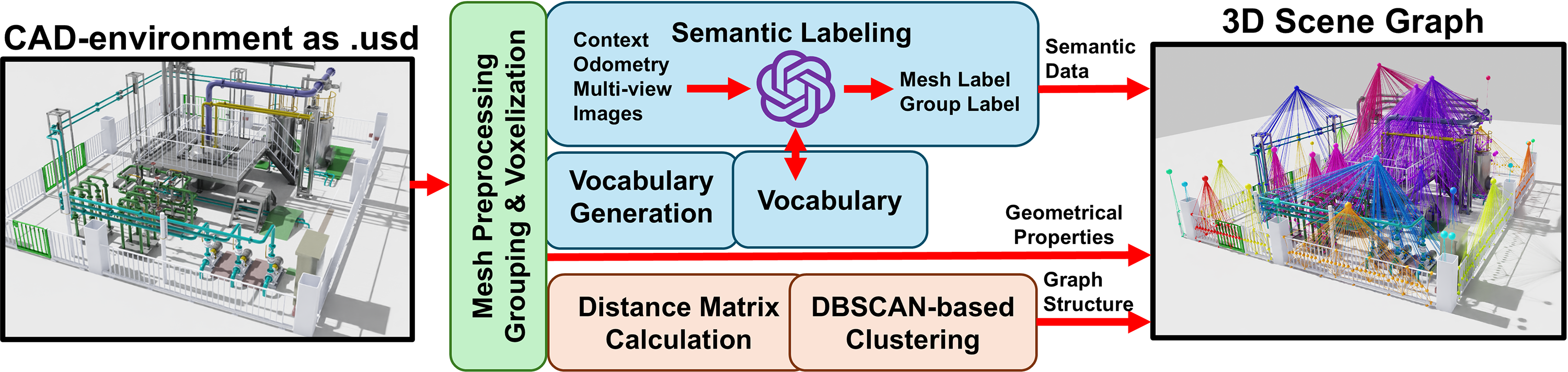}
    \caption{Overview of the process of generating a 3D scene graph from a CAD environment.}
    \label{fig:overview}
\end{figure*}


Industrial environments inherently pose various safety risks despite all precautionary measures. Thus, human workers are continuously exposed to hazardous situations. The deployment of robots offers a promising solution for taking over complex tasks and introducing advanced safety measures \cite{Trevelyan2016}. Prior to that, accurate industrial simulation environments are necessary for robot development, which are increasingly employed in research to develop and validate robotic perception, decision-making, task execution and robot-environment interaction under controlled and safe yet realistic conditions \cite{fernandez_perspective_2025,kargar_emerging_2024,choi_use_2021}. The effectiveness of these simulations is highly dependent on the richness and detail of the environment, particularly for tasks requiring high-level scene understanding.

Accurate industrial environments often come as CAD models, which provide detailed geometric and visual representations. However, several shortcomings, such as the lack of semantic annotations, relational information, functional insights and missing hierarchical organization, drastically limit the possibilities of high-level applications. While methods like SLAM and semantic mapping partially address these issues, generating 3D scene graphs provides a more comprehensive solution by creating a structured, holistic representation of an environment, including semantic, geometric and relational data \cite{hughes_hydra_2022, rosinol_kimera_2021}.

However, no published approaches to scene graph generation dealt with more complex environments such as factories or power plants. In such settings, crucial details are easily overlooked - for example, the segmentation of long-running, intertwined pipe networks. Furthermore, identifying and isolating key functional units, such as valves and gauges, can be difficult due to visual obstruction. Without structured, functional information, the simulation of dynamic elements remains out of reach for current approaches.

To address this gap, this paper contributes a novel and experimental approach to 3D scene graph generation from unstructured CAD models of complex industrial environments, as depicted in Fig. \ref{fig:overview}. Instead of relying on just a set of RGBD data forming a point cloud, we take an unstructured CAD file of the environment as a prerequisite. By employing a Large Visual-Language Model (LVLM), all meshes are individually semantically labeled and afterwards clustered regarding their minimal mesh distance. For clusters containing pipe structures, we analyze the scene graph to infer functional relations of gauges and valves inside the pipe systems. Thus, it can be captured how the objects functionally interrelate. 

Our results demonstrate that this approach successfully transforms a complex CAD environment into a multi-layered scene graph, preserving both the semantic and relational information of the environment. We show how this graph can be utilized to automatically extract functional units within a pipe system and identify their interconnections. This work provides a strong foundation for advanced applications in robotic scene understanding and the simulation of complex, dynamic industrial processes.

\subsection*{Main Contributions}
\begin{itemize}
    \item \textbf{A novel pipeline for scene graph generation from CAD models}, utilizing an LVLM for semantic enrichment and DBSCAN for spatial clustering to handle the complexity of industrial environments.
    \item \textbf{A multi-layered scene graph structure} that captures both low-level geometric relationships and high-level functional dependencies between components like pipes, gauges and valves, even in cluttered settings.
    \item \textbf{A method for automatically inferring functional relations within pipe systems} from the generated scene graph, enabling advanced scene reasoning and accessibility for LLM agents.
\end{itemize}

\section{Related Work}
The semantic understanding of 3D environments is a cornerstone of modern robotics, enabling high-level reasoning and task solving. Given raw data in the form of point clouds, semantic segmentation is often applied to extract objects from the collected point cloud \cite{rosinol_kimera_2021,SemSegInd2024,peng2023openscene}. 
LVLM such as gpt-4o are also being utilized for semantic labeling of objects in 3D environments \cite{li2025queryable}. It is shown that having multiple views on an object reduces mislabeling and can give further detailed descriptions \cite{li2025queryable}.

In addition, several works aim to build 3D scene graphs, organizing semantic and spatial information in an accessible format. Kimera generates layered dynamic scene graphs of indoor spaces using visual‑inertial SLAM and semantic segmentation \cite{rosinol_kimera_2021}. Focusing on the building's hierarchical structure, Armeni et al. structure 3D scene graphs over a building model and panoramic images \cite{armeni_3d_2019}. Hughes et al. propose a hierarchical real‑time representation for spatial perception, emphasizing scalability and loop closure support \cite{hughes_foundations_2024}. Wald et al. learn instance embeddings to predict object nodes and positional relations \cite{wald_learning_2022}. Scene graph creation can also be assigned to multiple separate actors, as in \cite{chang_hydra_multi_2023,greve_collaborative_2024}. Song et al. provide a comprehensive review of semantic mapping techniques, highlighting limitations of metric-only SLAM and the need for richer semantic structure \cite{song_semantic_mapping_2024}. Gan et al. focus on dynamic object scene graphs for long-term object search \cite{gan_longterm_search_2020}. 

Despite these advances, existing systems mostly rely on scene graphs generated from semantic point clouds for navigation or object search in indoor/domestic settings \cite{rosinol_3d_2020,rosinol_kimera_2021,hughes_hydra_2022,werby_hierarchical_2024}. Only \cite{armeni_3d_2019} requires a full environment model, yet, similar to the other approaches, industrial environments are not considered. Also, the lack of structure in such complex environments makes approaches like \cite{nguyen_translating_2024} not applicable, because objects are not named nor are they sorted in a meaningful way within the scene tree. Thus, functional relations such as in pipe systems cannot be identified with existing approaches. 
Having a 3D scene graph available is crucial to enable high-level tasks where an LLM interaction with the environment is necessary \cite{saxena2025grapheqa,li2025queryable}.

Our method uniquely integrates CAD input with LVLM-based mesh labeling, clustering of the environment's structures, and functional unit and relation identification, producing compact and functionally rich 3D scene graphs for industrial environments.

\section{Methodology}
The 3D scene graph is generated from a CAD environment model provided in the Universal Scene Descriptor (usd) file format. This model represents a single room (excluding walls and ceiling) from a robot test field location in Fukushima, Japan. We utilize Nvidia's Isaac Lab \cite{mittal2023orbit} to access the environment. The source model contains data about the geometry and colors/materials but is missing semantic labels and an insightful sorted structure of the meshes. For semantic labeling, OpenAI's gpt-4o is employed. The entire pipeline is programmed in Python using the Omniverse Python API \cite{omniverse}.

\subsection{Vocabulary} 
To ensure consistency throughout the semantic labeling process, a vocabulary of proposed labels is created beforehand. This vocabulary is organized in a three-layer tree structure, where the upper layer contains general terms that become more specific on the next layer (the root does not contain any information). For an excerpt of such vocabulary see the vocabulary box in Fig. \ref{fig:SemanticLabeling}. This structure allows each mesh to be assigned a general 'group' label and a specific individual 'name' label. The vocabulary is generated by providing gpt-4o multiple rendered images showing the environment from different angles sequentially. Based on the objects in the image, possible semantic labels will be identified and added to the vocabulary.

\subsection{Scene Graph}
The 3D scene graph is generated through the following steps: 
\begin{enumerate}
\item \textbf{Mesh Preprocessing:} Each mesh in the CAD environment is defined by its vertices and faces. Complex, curved shapes bring a high number of points with them, which increases computational load. Therefore, we voxelize all vertices onto a $1 cm$ grid, replacing each vertex with its nearest voxel center and subsequently delete duplicate points.

For distance calculation between meshes, the mesh faces are filled with points with $1 cm$ spacing along the straight lines between all three points that define a face. 

\item \textbf{Mesh Grouping:} CAD environments often contain an overwhelming amount of unsorted meshes of various sizes. To keep the amount manageable, we group very small meshes together with larger adjacent meshes. A threshold volume is defined for classifying small meshes. Consistent with the $1 cm$ voxel grid, we set the volume threshold to be $1cm^3$. Each mesh below this threshold is then assigned to the closest mesh above this threshold. If none is found in a defined proximity, the mesh is reclassified as a 'large' mesh. Henceforth, the term 'meshes' refers to these resulting mesh groups.

\item \textbf{Mesh Properties Embedding:} Each mesh is added as a node to the 3D scene graph. For visualization and further applications, geometric properties such as the mesh centroid and the 3D bounding box are stored as node attributes.

\begin{figure}
    \centering
    \includegraphics[width=0.5\textwidth]{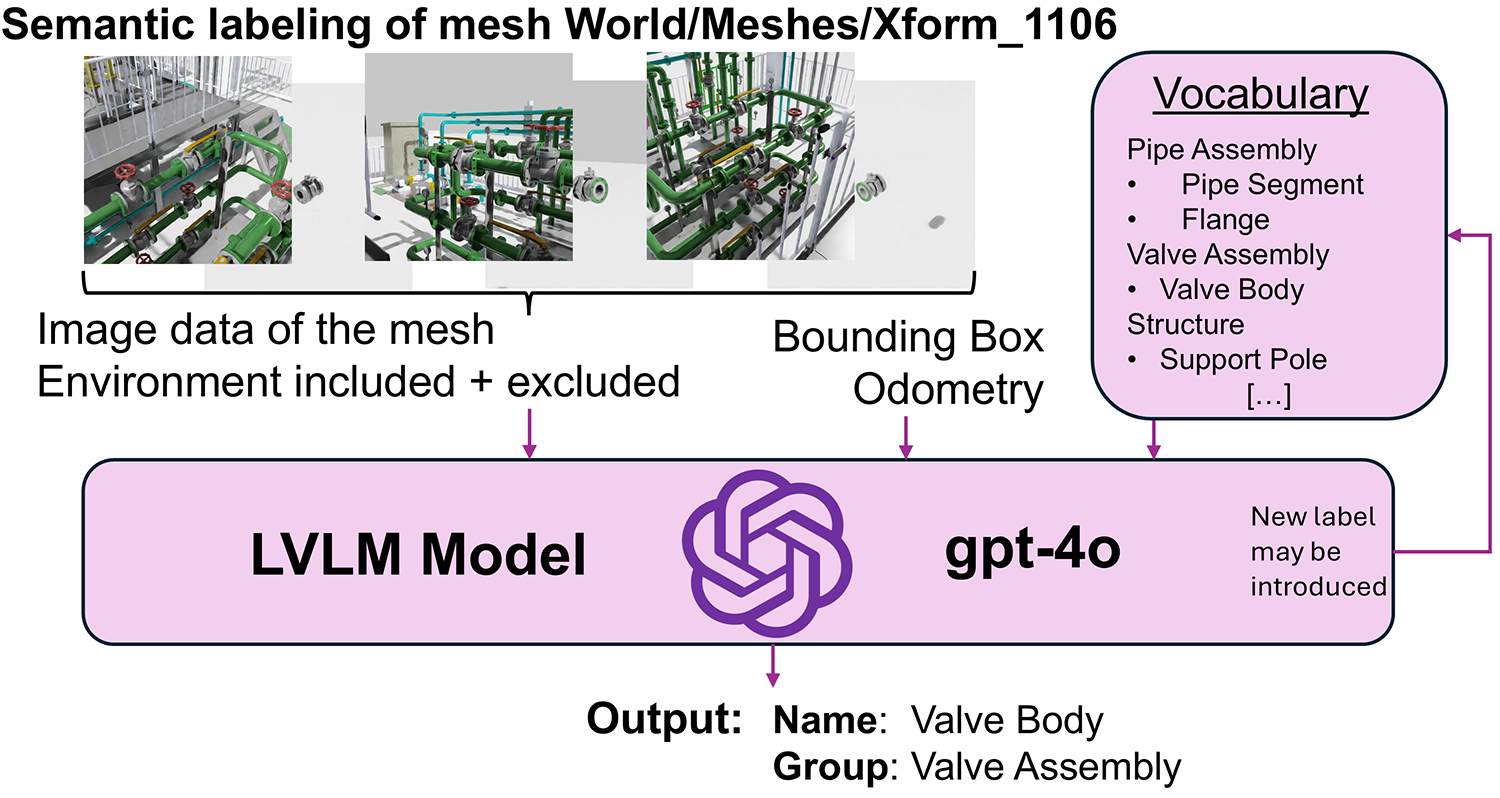}
    \caption{Visualization of the semantic labeling done for a selected mesh. The input consists of three pairs of two images, each from the same perspective, but one does include all other meshes of the environment and the other does not. Additionally, the bounding box sizes of the mesh and a vocabulary with pre-generated labels are given as input.}
    \label{fig:SemanticLabeling}
\end{figure}

\item \textbf{Semantic Labeling of Meshes:} We iterate over all meshes and use gpt-4o to assign semantic labels.  For each mesh, we take three pairs of two rendered RGB images (512x512 pixels). Each pair shows the mesh from a different camera view and consists of two images: 1) an image with all meshes visible and 2) an image with only the target mesh visible (illustrated in Fig. \ref{fig:SemanticLabeling}). Using these six images and the mesh's bounding box dimensions, gpt-4o describes the mesh using a 'group' and 'name' label from the vocabulary. If no suitable label exists, it is permitted to propose a new one. The resulting labels and the mesh's usd path are then added as attributes to the corresponding node.

\item \textbf{Clustering of Meshes:} To segment distinct structures, we use DBSCAN \cite{dbscan} to cluster the nodes by their minimal mesh points' proximity. We configure DBSCAN with $\epsilon = 0.01$ and $min\_samples = 1$. Due to the voxel grid size of $1 cm$, a smaller $\epsilon$ would lead to over-fragmented clusters and a larger $\epsilon$ would consider not touching meshes also as meshes. The ground plane is excluded from this process. Based on the clustering results, edges representing spatial adjacency are added between two nodes if the distance is not larger than $1 cm$ and only within the same cluster. Finally, we add a new parent node for each cluster with edges to all nodes of its cluster. 

\item \textbf{Identifying Functional Relations:} For clusters with pipe systems, functional relations can now be analyzed. Using the scene graph and the 'group' semantics, Algorithm \ref{alg:funcRel} infers the direct functional relationships between the functional units inside the pipe system. This algorithm requires a predefined list of "connector" labels (here 'Pipe assembly'), denoted $S_{\text{pipe}}$, allowing it to traverse the graph while ignoring non-relevant meshes like structural supports. A list of functional units is also necessary, which is obtained by finding interconnected node clusters of the same 'group' label.
\begin{algorithm}
\caption{Extracting Functional Relations}
\label{alg:funcRel}
\begin{algorithmic}[1]
\Require
    Scene Graph $\mathcal{G} = (\mathcal{V}, \mathcal{E})$;
    Semantic map $S: \mathcal{V} \to \mathcal{L}$, where $\mathcal{L}$ is the set of 'group' labels;
    Set of pipe semantic labels $\mathcal{S}_{\text{pipe}} \subseteq \mathcal{L}$;
    Set of functional units $\mathcal{F} = \{F_1, F_2, ..., F_k\}$, where each $F_i \subseteq \mathcal{V}$

\State $M \leftarrow \bigcup_{F_i \in \mathcal{F}} F_i$ \Comment{Initialize global marked set with all nodes of functional units}
\While{$M$ changed in size}
    \For{each functional unit $F_i \in \mathcal{F}$}
        \State $L_{\text{new}} \leftarrow \emptyset$ \Comment{Nodes to add to $F_i$ in this pass}
        \For{each node $v \in F_i$}
            \For{each neighbor $u$ of $v$ in $\mathcal{G}$}
                \If{$u \notin M$ \textbf{and} $S(u) \in \mathcal{S}_{\text{pipe}}$}
                    \State $L_{\text{new}} \leftarrow L_{\text{new}} \cup \{u\}$
                    \State $M \leftarrow M \cup \{u\}$ \Comment{Claim $u$ for $F_i$}
                \EndIf
            \EndFor
        \EndFor
        \State $F_i \leftarrow F_i \cup L_{\text{new}}$ \Comment{Expand the functional unit with new neighbouring nodes}
    \EndFor
\EndWhile

\State $\mathcal{V}_{\text{func}} \leftarrow \{1, 2, ..., k\}$ \Comment{functional units named after indices of $F_i$}
\State $\mathcal{E}_{\text{func}} \leftarrow \emptyset$
\For{each edge $(v, u) \in \mathcal{E}$}
    \State Find $i, j$ such that $v \in F_i$ and $u \in F_j$
    \If{$i$ and $j$ exist \textbf{and} $i \neq j$}
        \State $\mathcal{E}_{\text{func}} \leftarrow \mathcal{E}_{\text{func}} \cup \{(i, j)\}$
    \EndIf
\EndFor
\State $\mathcal{G}_{\text{func}} = (\mathcal{V}_{\text{func}}, \mathcal{E}_{\text{func}})$
\State \Return $\mathcal{G}_{\text{func}}$
\end{algorithmic}
\end{algorithm}

\begin{figure}[]
\centering
\includegraphics[width=0.5\textwidth]{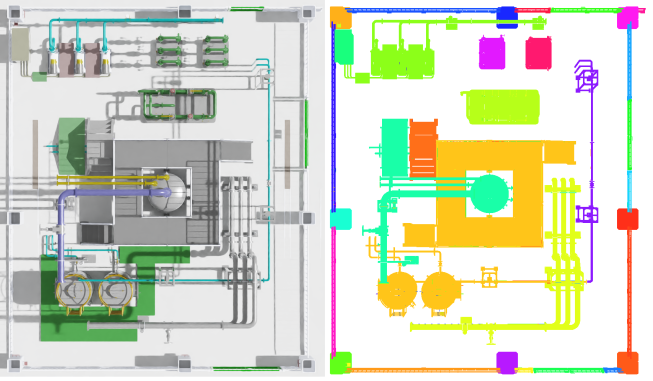}
\caption{Results after applying DBSCAN to the environment shown left with parameters $\epsilon = 0.01$ and $min\_samples = 1$. }
\label{Fig:SG}
\end{figure}
\end{enumerate}

\section{Results}
The test environment (see first image in Fig. \ref{fig:overview}) comprises 8,327 meshes. After excluding 9 meshes identified as modeling artifacts, the grouping process (merging meshes $\leq 1cm^3$ with larger neighbors) reduced the amount to 2,068 meshes. 
\subsection{Clustering Results}
Applying DBSCAN with $\epsilon=0.01$ and $min\_samples = 1$ resulted in 39 clusters. Excluding the outer fences, we are left with 15 clustered structures. The clustering is visualized in Fig. \ref{Fig:SG}. The algorithm successfully isolated freestanding structures and correctly grouped complex multi-pipe systems. However, we observed two main clustering errors: (1) one pipe structure was incorrectly split into two clusters and (2) the central access platform was clustered together with a pair of tanks, despite having no visible physical connection.


\begin{figure*}
\centering
\includegraphics[width=\linewidth]{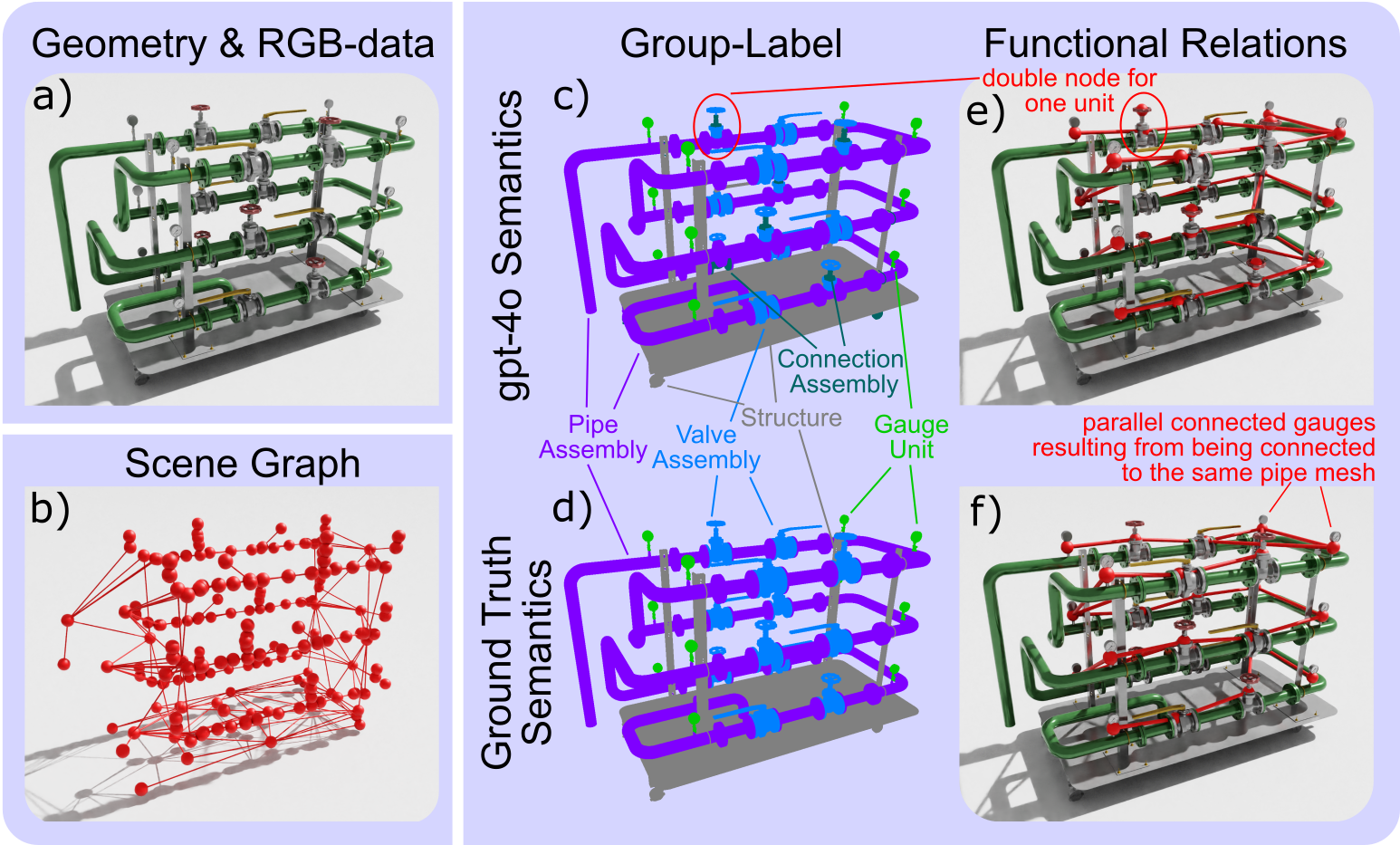}
\caption{Visualization of the scene graph, semantic data and functional analysis pipeline for a selected pipe structure. a) The original rendered geometry. b) The generated spatial scene graph, where nodes (spheres) represent meshes and edges indicate spatial proximity. c) Semantic 'group' labels assigned by gpt-4o. d) The manually assigned ground truth labels. e) The extracted functional graph (functional units shown in red) derived from the LVLM-generated semantics in c). f) The extracted functional graph derived from the ground truth semantics in d).}
\label{Fig:PipeResult}    
\end{figure*}

\begin{figure}
\centering
\includegraphics[width=0.5\textwidth]{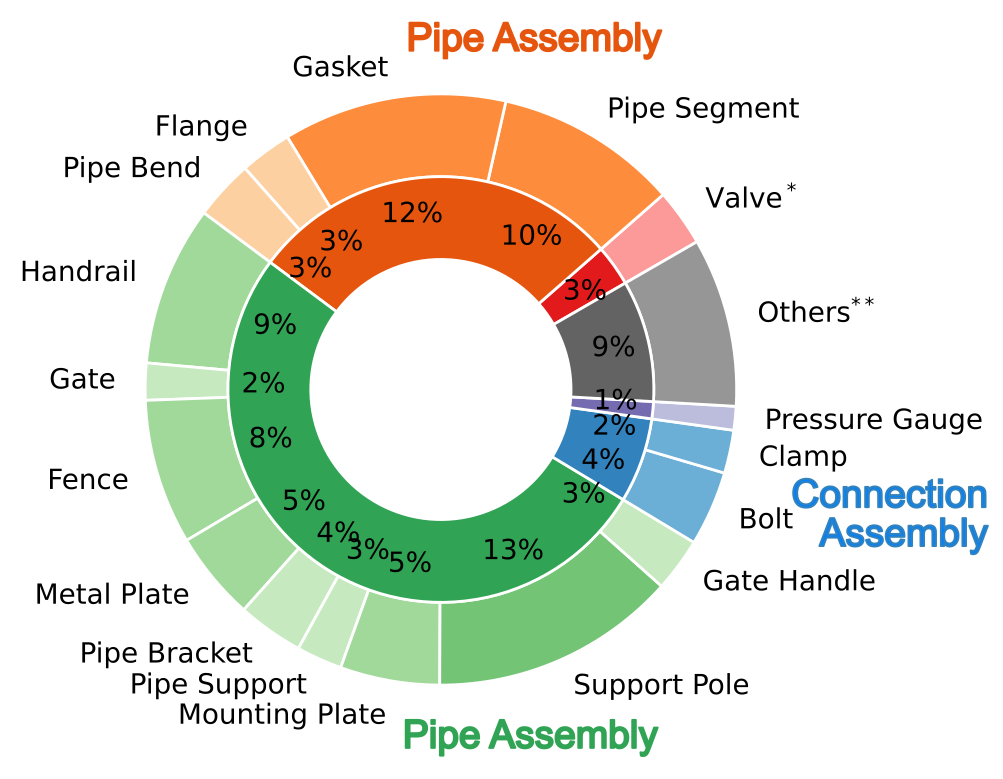}
\caption{Distribution of generated semantic labels in the used CAD environment. Similar colors indicate that the semantic group label is equal. Used group labels but not separately visualized are 'Pump Unit' and 'Connection Assembly'. The amount of labeled meshes is 2068.
\\
{\normalfont\footnotesize * Valve combines all valve-related semantic labels. \\ ** Others contain semantic labels with a quantity less than 25.}}
\label{Fig:Pie}
\end{figure}

\begin{figure}
\centering
\includegraphics[width=0.5\textwidth]{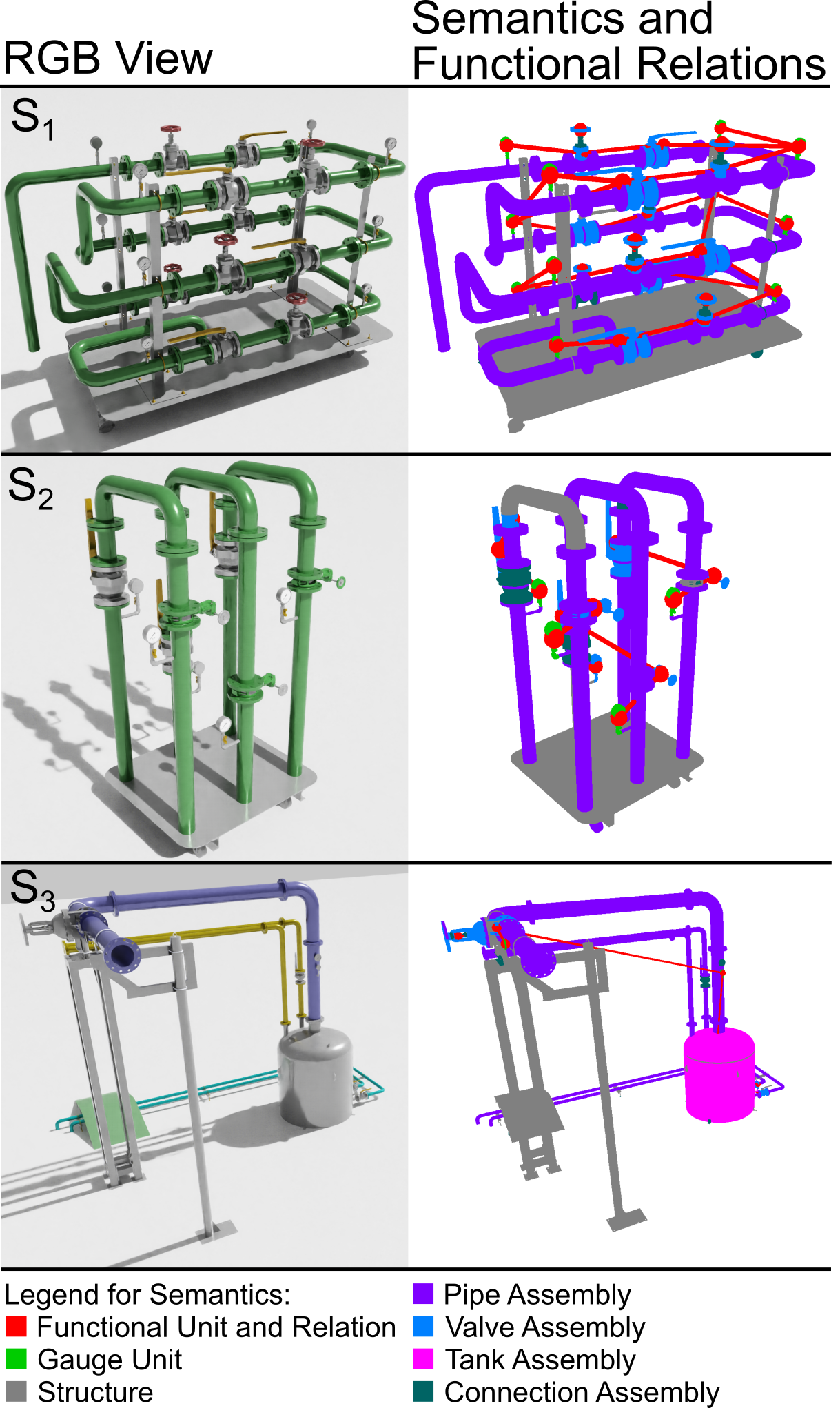}
\caption{Three selected structures, where each is a clustering result of using DBSCAN. The left side shows the structures in normal sight. The right side shows the semantic 'group' label and identified functional relations.}
\label{Fig:PipeStructuresResult}    
\end{figure}

\subsection{Clustering Labeling Results}
Each of the 2,068 meshes was assigned two semantic labels. A total of 7 'group' labels and 44 distinct 'name' labels were used in the final labeling. In the final label distribution (see Fig. \ref{Fig:Pie}), meshes corresponding to structural elements constitute the majority of the distribution. Pipe-related meshes make up around one quarter. Valves were either labeled as such or were put under the label of 'connection assembly'. 

The semantic accuracy was evaluated for three selected pipe structures (also classified as single clusters by DBSCAN), with quantitative results presented in Table \ref{tab:SemanticAccuracy}. A ground truth was manually created using the same vocabulary. This ground truth vocabulary was expanded with new labels where necessary, a process our LVLM-based method can also perform.

\begin{table}
    \caption{Semantic accuracy evaluated on three structures inside the environment, which are shown in Fig. \ref{Fig:PipeStructuresResult}.}
    \begin{tabular}{r|c|c|c}
         & $S_1$ & $S_2$ & $S_3$ \\ \hline
        Semantic Mesh Label Accuracy & $38.5\%$  & $30.5\%$  & $42.0\%$   \\
        Semantic Group Label Accuracy & $79.9\%$  & $73.2\%$  & $84.0\%$ \\
        Total amount of Meshes & $174$  & $82$  &  $200$ \\
        &   &   &   \\
        Fully Found Valves & $5$  & $2$  &  $0$ \\
        Partially Found Valves & $7$  & $4$  &  $3$ \\
        Missed Valves & $0$  & $0$  &  $1$ \\
        Fully Found Gauges & $12$  & $4$  &  $3$ \\
        Missed Gauges & $0$  & $2$  &  $2$ \\
    \end{tabular}
    \label{tab:SemanticAccuracy}
\end{table}

\subsection{Functional Relation Analysis}
For a more qualitative analysis, we examine the relations between functional units of a pipe system. Fig. \ref{Fig:PipeStructuresResult} visualizes identified functional relations of three semantically evaluated structures of Table \ref{tab:SemanticAccuracy}. 
\begin{itemize}
    \item \textbf{Structure $S_1$}, contrasts in Fig. \ref{Fig:PipeResult} the results of using ground truth semantics (Fig. \ref{Fig:PipeResult}d) versus gpt-4o-generated semantics (Fig. \ref{Fig:PipeResult}c). While the ground truth semantics lead to all functional units being identified (Fig. \ref{Fig:PipeResult}f), the gpt-4o-based semantics caused wheel valves to be incorrectly identified as two separate functional units. Furthermore, both semantics showed uncertainty regarding the order of the gauges, which are two each connected to the same pipe mesh
    \item For \textbf{structure $S_2$} all functional relations except two missing gauges were identified correctly and four out of six valves were only partially labeled as valves.
    \item \textbf{Structure $S_3$} features a central tank that correctly links multiple pipe systems into a single functional graph. However, two gauges were again missed, one of which one was misidentified as a valve. Additionally, one valve was completely mislabeled as 'connection assembly' and hence was not considered as a functional unit.
\end{itemize}

\section{Discussion}
The results demonstrate that our approach can generate a semantically rich, functional 3D scene graph, but they also highlight several challenges of industrial environments that merit further discussion.

\subsection{CAD Meshes vs. Point Cloud Processing}
A key challenge in industrial scene understanding is the high level of detail, which contrasts sharply with typical office or home environments. SLAM and semantic mapping techniques, which operate on point clouds, typically focus on coarse object detection (e.g., chairs, monitors) \cite{rosinol_kimera_2021 ,werby_hierarchical_2024}. While industrial-focused instance segmentation \cite{agapaki2021instance} exists, it often involves high computational demands. Additionally, these approaches cannot disassemble assemblies into their various components, as scanned point clouds give only visible surface information about the object.

Our approach bypasses these segmentation challenges by operating directly on the CAD-defined meshes. This provides inherent, high-fidelity object granularity from the outset. However, CAD models introduce their own challenges. The environment contained several redundant or non-visible meshes and a high number of small detailed components (e.g., bolts, modeling artifacts). This strongly affects the computation time of DBSCAN, which requires a full distance matrix of all meshes, making a preprocessing step crucial. Aggregating meshes smaller than 1 cm³ reduced the total mesh count by approximately $75\%$ and thereby significantly decreased computational load and LVLM prompting costs.

The $1cm$ voxelization was essential for tractability but also introduced two clustering errors: splitting of a continuous pipe into two distinct clusters and clustering two unrelated structures together (caused by an unexpected single edge connecting them). This suggests that while proximity-based clustering is effective in separating most intertwined structures, it is sensitive to voxelization artifacts and may require more geometry awareness during the preprocessing step. 

\subsection{Semantic Labeling}
The semantic distribution in Fig. \ref{Fig:Pie} indicates that a significant portion of meshes can be classified into a finite number of distinct object categories, supporting scene understanding. However, the ratio of gaskets to flanges is logically inconsistent. As one gasket should be in between two flanges, a 1:2 ratio would be expected, but the results show a strong over-prediction of gaskets. It is suggested that gpt-4o has misclassified many flanges as gaskets, even though the bounding box dimensions were provided that should have distinguished the thin gaskets from the thicker flanges. Consequently, it is likely that other technical labels may also be incorrect. 

This indicates that our approach to semantic labeling, which uses three RGB images pairs and bounding box data, is insufficient for fine-grained technical object recognition. Improvements could be achieved by matching against a CAD model catalog, which is already used for semantic segmentation \cite{rosinol_3d_2020,armeni_3d_2019}. CAD models for technical components could be provided from sources such as \cite{GrabCAD}. However, such foreign CAD models might not align with the objects in our environment. An alternative to using external catalogs would be to leverage the environment's own meshes, identifying component repetition to enforce labeling consistency.

\subsection{Vocabulary}
The use of a semantic vocabulary helps maintain label consistency, which is crucial for subsequent analysis. While this risks a minor loss of information, as the LVLM tends to reuse existing labels rather than create new ones, it results in a more structured output. In combination with the scene graph, the semantic labels get more expressive through contextual information from neighboring nodes. Without an initial vocabulary, the label diversity can vary strongly. Therefore, it is crucial to predefine certain groups of interest. Less critical groups, such as structural ones, can be intentionally neglected.

\subsection{Functional Relation Analysis}
A primary goal of this work is to enable high-level reasoning by extracting functional relations. As shown in Figs. \ref{Fig:PipeResult} and \ref{Fig:PipeStructuresResult}, our pipeline successfully extracts functional systems from the scene graph. The accuracy of the semantic information as well as a correct scene graph are essential for successful identification and determining the right order. Nonetheless, the comparison for structure $S_1$ (see Fig. \ref{Fig:PipeResult}) shows that even with partially incorrect semantics (e.g., a valve assembly being only partially labeled as one), the position and sequence of functional relations was correctly identified, demonstrating robustness to minor labeling errors. Still, it may lead to unconnected artifacts of functional units, as wheel valves, for example, tend to be split into two separate valve nodes because an intermediate mesh was mislabeled as a 'connection assembly' and not a 'valve assembly'. 

However, it is observed that Algorithm \ref{alg:funcRel} has certain analysis limitations.
First, it cannot resolve the order of functional units connected solely to one common mesh (e.g., two gauges on one pipe segment as in $S_1$). The scene graph's topology is insufficient here and more detailed geometric calculations would be needed to determine the exact order.
Second, the scene graph captures only connectivity but not directionality; for the tank in structure $S_3$, it remains ambiguous which pipes are inputs and which are outputs. 
Finally, our environment did not feature complex branching pipes, leaving the applicability here open to future research. 

Our scene graph, enriched with these functional relations, provides the structured representation required for reasoned simulation of dynamic elements. LLM models can be used for reasoning over structured data \cite{jiang-etal-2023-structgpt} and are also already utilized with scene graphs for task planning and scene analysis \cite{werby_hierarchical_2024,li2025queryable,saxena2025grapheqa}. Hence, giving an LLM the scene graph including functional relations, synthetic simulation data can be generated for use in the environment, enhancing the dynamic of simulations.

\subsection{Limitations}
Despite achieving good and insightful results, several limitations are left to address in future work. 
First, the preprocessing steps required for scalability introduce geometric artifacts. Voxelization reduces the complexity effectively but can lead to clustering errors of the complex environment. However, tackling larger or more detailed environments would stress the scalability issue of especially DBSCAN. 

Second, the semantic labeling accuracy for specialized industrial components remains a key challenge. The general-purpose LVLM struggled with fine-grained distinctions, leading to systematic misclassifications. These semantic errors propagate directly into the functional analysis, creating incomplete or perhaps false functional units.

Finally, the functional relation extraction can be ambiguous in complex cases due to the nature of the environment model.

\section{Conclusion}
This paper presents a novel approach to generating multi-layered 3D scene graphs from an industrial simulation environment available as a CAD file. As the environment consists solely of geometric and RGB data, it is difficult to embed high-level applications, especially if an interaction of an LLM agent or robot with the environment is required. 

Our resulting 3D scene graph provides a semantically rich, structured representation of an industrial facility. The graph embeds geometric properties such as position and size, while its edges represent spatial adjacency. This hierarchical structure offers a compact yet expressive abstraction of the environment's complex geometry.

Furthermore, this scene graph can be utilized to identify functional units such as gauges and valves within pipe systems. The functional relation of those can then be extracted, giving insight about which components influence what other components and in what order. This sets the necessary foundation for creating dynamic simulations of functional units within the simulation environment that test a robot's capability to understand and solve high-level problems in a complex industrial environment. 





\bibliographystyle{IEEEtran}
\bibliography{IEEEabrv,root}

%
%
%

\end{document}